\documentclass[11pt]{article}

\usepackage[preprint]{acl}

\usepackage{times}
\usepackage{latexsym}

\usepackage[T1]{fontenc}

\usepackage[utf8]{inputenc}

\usepackage{microtype}

\usepackage{inconsolata}

\usepackage{graphicx}
\usepackage{booktabs}

\usepackage{tcolorbox}

\title{Developmental Trajectories of Situation Modeling and Mentalizing in Transformer Language Models}

\author{
  Pamela D. Rivière \\
  Rutgers University - Newark \\
  \texttt{pamela.riviereruiz@rutgers.edu}
  \And
  Cameron Jones \\
  Stony Brook University \\
  \texttt{cameron.jones@stonybrook.edu}
  \AND
  Sean Trott \\
  Rutgers University - Newark \\
  \texttt{sean.trott@rutgers.edu}
}
  
\begin{document}
\maketitle
\begin{abstract}
Recent work suggests that Large Language Models (LLMs) are sensitive to the belief states of agents described by text, as measured by the false belief task (FBT), yet persistent concerns of construct validity remain. We adopt a \textit{developmental perspective}, tracing the pattern of mental state reasoning behavior---and likely \textit{preconditions} for this behavior---across multiple training stages in the Olmo$2$ and Pythia language model suites. We find that above-chance FBT performance depends both on model size and sufficient training volume, emerges relatively late in pretraining, and is most improved by post-training interventions  (SFT, DPO) in the condition most diagnostic of mentalizing (False Belief, Implicit). However, FBT performance is fragile: consistent with past work, the use of non-factive verbs (e.g., ``thinks'') increases false belief attributions even in the True Belief condition. To contextualize these findings, we track the emergence of \textit{situation modeling}: the ability to report on basic factual properties of a described scene. Situation modeling accuracy generally precedes and exceeds FBT accuracy, yet situational representations also prove surprisingly incoherent in certain respects: when asked about the knowledge states of the \textit{Antagonist} agent---who always knows the item's true location---Olmo$2$ 13$b$ is consistently influenced both by the \textit{Target} agent's knowledge state and the presence of non-factive verbs. Together, these results suggest that larger, sufficiently trained models build partially coherent situation models in a developmentally appropriate sequence, yet display surprising fragility---highlighting the value of developmental and stress-testing approaches for evaluating LLM capabilities.
\end{abstract}

\section{Introduction}\label{sec:intro}

The capacity to infer and reason about the belief states of others---sometimes called \textit{mentalizing}---serves as a powerful cognitive tool \citep{apperly_what_2012}. It plays a crucial role in interpreting and predicting the behavior of other social agents \citep{apperly_what_2012}, and may facilitate cooperation \citep{tomasello_understanding_2005} or even persuasion \citep{slaughter_i_2013}. Past research has explored whether and when this capacity develops in children \citep{wellman_meta-analysis_2001}, non-human animals \citep{krupenye_great_2016}, and more recently, large language models (LLMs) \citep{kouwenhoven2026social, trott_large_2023, shapira_clever_2024, ullman_large_2023, kosinski_evaluating_2024, xu_opentom_2024, gandhi_understanding_2023, jones_comparing_2024, jones_does_2024, street_llms_2026}.

The question of whether LLMs can \textit{mentalize} has generated considerable excitement. It informs longstanding hypotheses about the role of language input in shaping human mentalizing capacity \citep{brown_why_1996, de_villiers_role_2014, trott_language_2026}. Practically, the design and interpretation of mentalizing ``benchmarks'' for LLMs \citep{xu_opentom_2024, jones_comparing_2024, gandhi_understanding_2023} could affect decisions about how to interpret or control downstream behaviors of concern, such as deception or persuasion \citep{jones_lies_2026}. Yet both aims are complicated by persistent concerns about \textit{construct validity} \citep{trott_large_2023, hu_re-evaluating_2025, shapira_clever_2024, ullman_large_2023, ivanova_how_2025}: do tests of mentalizing in LLMs actually assess the construct they purport to measure---and how would we know? 

Indeed, construct validity concerns pervade LLM benchmarking practices \citep{raji_ai_2021, saxon_benchmarks_2024, bean_measuring_2025}, research on human cognition and behavior \citep{borsboom_concept_2004, bloom_two_2000}, and the history of scientific measurement more broadly \citep{chang_inventing_2004}. Researchers typically address validity concerns using various techniques, including analyses of convergent and predictive validity \citep{alexandrova_is_2016}. 

In the current work, we adopt a complementary perspective that emphasizes the \textit{developmental trajectory} of an LLM's mentalizing behavior throughout pretraining. In both human psychology and the study of LLM behavior, a developmental perspective has generated fruitful insights into not only the \textit{final state} of a system's behavior but the key mechanisms and milestones that precede it \citep{chen_sudden_2023, tigges_llm_2024, riviere2025start}; see also Section \ref{subsec:developmental}. 

Specifically, a developmental perspective allows us to ask several targeted research questions. First, \textit{when} during pretraining or post-training does putative mentalizing behavior ``emerge'' (Section \ref{sec:mentalizing-results})? Second, how does the emergence of mentalizing behavior relate to other capacities that are arguably \textit{preconditions} to the ability to reason about belief states, e.g., the ability to track basic properties of a \textit{situation} (Section \ref{subsec:sm_results})? This has analogies to developmental work more broadly \citep{chen_sudden_2023} and also serves the function of ``stress-testing'' measures of LLM capabilities \citep{riviere2025start, shapira_clever_2024, ullman_large_2023}. Arguably, the extent to which mentalizing-like behavior is stable, robust to stress-testing, and preceded by behavioral preconditions should shape our intuition about whether that behavior reflects mentalizing per se or a more disparate collection of shallow, task-dependent heuristics \citep{shapira_clever_2024}.  

The remainder of this paper is structured as follows: first, in Section \ref{sec:related}, we describe related work on mentalizing in humans and LLMs, construct validation, and the value of a developmental perspective. In Section \ref{sec:methods}, we describe our methods and the rationale for our experimental design and analyses. We then present the results of our analysis of the developmental trajectory of mentalizing beahvior across select language models (Section \ref{sec:mentalizing-results}) and its relation to other behaviors, such as \textit{situation modeling} (Section \ref{subsec:sm_results}). Finally, we discuss the implications of this work for the study and interpretation of mentalizing behavior in LLMs, as well as work on identifying LLM ``capabilities'' more generally (Section \ref{sec:discussion}).

\section{Related Work}\label{sec:related}

\subsection{Mentalizing in humans and LLMs}\label{subsec:mentalizing}

As described in Section \ref{sec:intro}, \textit{mentalizing} has long been a topic of interest throughout Psychology, and more recently, in the study of LLM behavior or ``machine cognition''. This capacity is often assessed using a variant of the \textit{false belief task} (FBT) \citep{wimmer1983beliefs, baron1985does, krupenye_great_2016}. In the typical design, a participant witnesses an object being moved from one location to another, with or without another person's knowledge. The participant is then asked about that person's likely belief states, either explicitly
(e.g., ``Where does Sally think the book is?'') or implicitly (e.g., by predicting where Sally might look). Successful performance requires \textit{sensitivity} to the other person's implied knowledge states---responding
differently depending on what that person likely knows or believes.

The FBT is widely used as an index of sensitivity to mental states \citep{bradford2020neural, schneider2014implicit, papafragou2025pragmatic}, and has been adapted to the study of mentalizing behavior in non-human animals \citep{krupenye_great_2016} and LLMs \citep{kouwenhoven2026social, strachan_testing_2024, street_llms_2026, kosinski_evaluating_2024, trott_large_2023}. That said, it has faced a number of criticisms over the years, including the concern that it indexes additional capacities beyond mentalizing \citep{bloom_two_2000} and the fact that it displays limited convergent and predictive validity in humans \citep{gernsbacher2019empirical, hayward2017reliability}. This has led some to call into question the coherence of ``Theory of Mind'' or ``mentalizing'' as a construct altogether \citep{gough2023does}, presaging analogous debates about measures of mentalizing in LLMs \citep{hu_re-evaluating_2025} and prompting the creation of new tasks \citep{buttelmann2014eighteen, dodell2013using, trott2019individual, jones_comparing_2024}. 

We are not committed here to a strong view regarding the validity of the FBT as a definitive window into mentalizing capacities more broadly. Our interest, rather, is in the FBT as a behavioral index of \textit{mental state sensitivity}. This index can then be traced developmentally and also compared to other behavioral indices, such as the ability to accurately report on the basic features of a situation (see Section \ref{subsec:sm_results}). Indeed, these analyses might inform questions about the validity of the FBT for LLMs, roughly analogous to the logic underlying convergent validity \citep{alexandrova_is_2016}.

\subsection{A Developmental Perspective}\label{subsec:developmental}

The study of development or \textit{ontogeny} can reveal unique insights that are more challenging to attain from a cross-sectional perspective. Research on human cognition has long incorporated a developmental focus, from the early work of Piaget \citep{piaget_construction_1954} to more contemporary work on child development \citep{gopnik1999scientist, tomasello2016early}, including work on mentalizing specifically \citep{wellman_meta-analysis_2001}. One key scientific benefit of this perspective is that it allows researchers to observe how a particular behavior unfolds across \textit{time}, addressing questions about whether this pattern consists of discrete ``jumps'' or continual improvement \citep{wellman_meta-analysis_2001}, as well as what kinds of experiential input are necessary or sufficient for the emergence of a particular behavior \citep{astington1999longitudinal}. Similarly, researchers can ask about the \textit{temporal ordering} of different behaviors, granting insight into the requisite capacities or preconditions facilitating a particular capacity of interest \citep{de2016sensorimotor}. 

More recently, research on LLMs has also adopted a developmental perspective, with the goal both of learning more about human developmental, e.g., language learning \citep{warstadt-etal-2023-findings, constantinescu2025investigating} and also learning more about LLMs themselves \citep{chen_sudden_2023, saphra-lopez-2019-understanding, riviere2025start, michaelov2025language, zhao2024distributional}. This approach carries many of the same advantages described above, e.g., investigating which behaviors are learned in which order \citep{chen_sudden_2023}, and it also allows researchers to connect emergent behaviors to low-level mechanisms or ``circuits'' \citep{olsson2022context, tigges_llm_2024, riviere2025start, trott_toward_2025, aoyama_predicting_2025}. 

Finally, concurrent work \citep{kouwenhoven2026social} conducted a case study on the developmental pattern of false belief reasoning in Olmo2 \citep{olmo20242}. Two key results are relevant here: first, syntactic competence emerges much earlier during pretraining than competence on the FBT; and second, post-training had nuanced effects on FBT performance, with small improvements observed for the \textit{false belief} condition but more mixed effects on the \textit{true belief} condition. We report similar results in Section \ref{sec:mentalizing-results} (see \textbf{Figure \ref{fig:acc-post-train}}). 

The current work extends this line of inquiry in two key ways. First, by comparing the OLMo~2 suite with the (relatively) undertrained Pythia suite \citep{biderman2023pythia}, we examine the \textit{joint} contribution of model size and volume of language exposure---finding that neither alone is sufficient. Second, and more critically, we investigate whether putative mentalizing behavior is preceded by the emergence of a more basic capacity: the ability to accurately track descriptive properties of a situation, such as where an object is located
or who moved it (Section~\ref{subsec:sm_results}). This \textit{situation modeling} capacity represents a task-specific precondition to false belief reasoning, complementing
\citeauthor{kouwenhoven2026social}'s comparison of FBT performance with general syntactic competence.

\section{Materials and Methods}\label{sec:methods}

\subsection{Mentalizing Stimuli}
To examine language model (LM) mentalizing ability, we used a publicly accessible False Belief Task (FBT) stimulus set \citep{trott_large_2023}. Stimuli consist of passages describing scenes in which two different named agents serially change the location of an item such that its starting and final locations differ. Crucially, a target agent has either observed or failed to observe the item's relocation. Each passage ends with the implication or indication that this target agent will subsequently attempt to retrieve the item, and the LM's task is to produce probabilities over Start and End locations, indicating where the agent is most likely to search. 

\citet{trott_large_2023} created a total of $192$ unique passages from $12$ templates by varying properties that might influence performance on the FBT, such as \texttt{Knowledge State} (True or False Belief; reflecting whether the target agent observes the item's displacement to its final location or not) and \texttt{Knowledge Cue} (Explicit or Implicit; whether the final sentence references the agent's beliefs (e.g. ``Dave \textbf{thinks} the book is in the...'' or ``Dave \textbf{looks for} the book in the...''). Start and End locations are each mentioned twice per passage; passages consequently vary to cover all permutations of the locations' first and last mentions. Sample passages can be found in Appendix \ref{subsec:app-stims}.

\subsection{Situation Modeling Stimuli}\label{sec:sm-stims}
To further stress-test \citep{shapira_clever_2024,riviere2025start} LMs' abilities to create and sustain a Situation Model---that is, to form coherent representations of agents and events---we modified the original stimulus set to obtain LM probabilities that should reflect descriptive properties of a passage which the LM need only egocentrically report on. In principle, such a capacity is likely to serve as a crucial prerequisite to the more complex ability to attribute mental states to agents who have beliefs that diverge from an omniscient observer's. 

\begin{figure}[t]
  \includegraphics[width=\columnwidth]{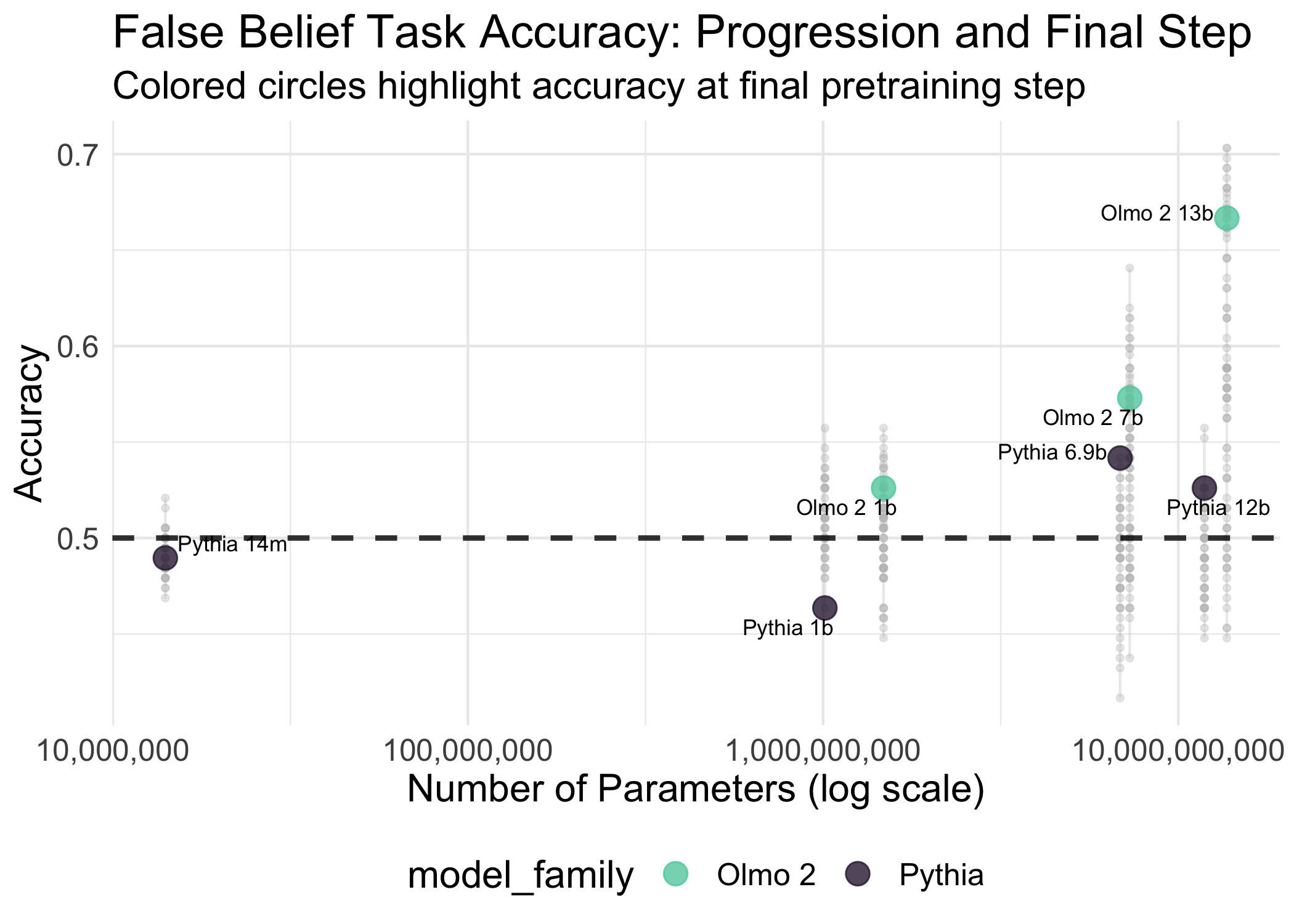}
  \caption{False belief task accuracy increases as a function of LM size for the Olmo$2$ suite. Grey observations correspond to a given LM's performance at distinct pretraining checkpoints. Observations colored according to model family correspond to an LM's performance at its final pretraining step (for Olmo$2$, this is the final step in stage$1$ pretraining). Horizontal dashed line marks chance performance.}
  \label{fig:acc-by-model-size}
\end{figure}

We first tested LMs' ability to track the belief states of the Antagonist, who always moved the item to its end location. These $192$ passages were identical to the original set, except for the named agent in the final critical sentence (example in Appendix \ref{sec:appendix}). In all of these cases, if the LM is capable of constructing an accurate Situation Model, it should produce larger probabilities for End locations than for Start.

For the second block of stress-tests, we took each of the $192$ passages and added one of four new query sentences whose best completions were the item's actual location, or the name of the agent responsible for having placed the item in either its Start or End locations ($n = 768$ passages). Examples below:
\begin{itemize}
  \item At the start of the story, John put the shovel in the
  \item At the end of the story, the shovel was in the
  \item The person who put the shovel in the toolbox was
  \item The person who moved the shovel to the van was
\end{itemize}

Queries were designed to match the way that information was presented in the passage. Appendix \ref{sec:appendix} additionally documents performance on minimal-pair versions of these queries.

\begin{figure}[t]
  \includegraphics[width=\columnwidth]{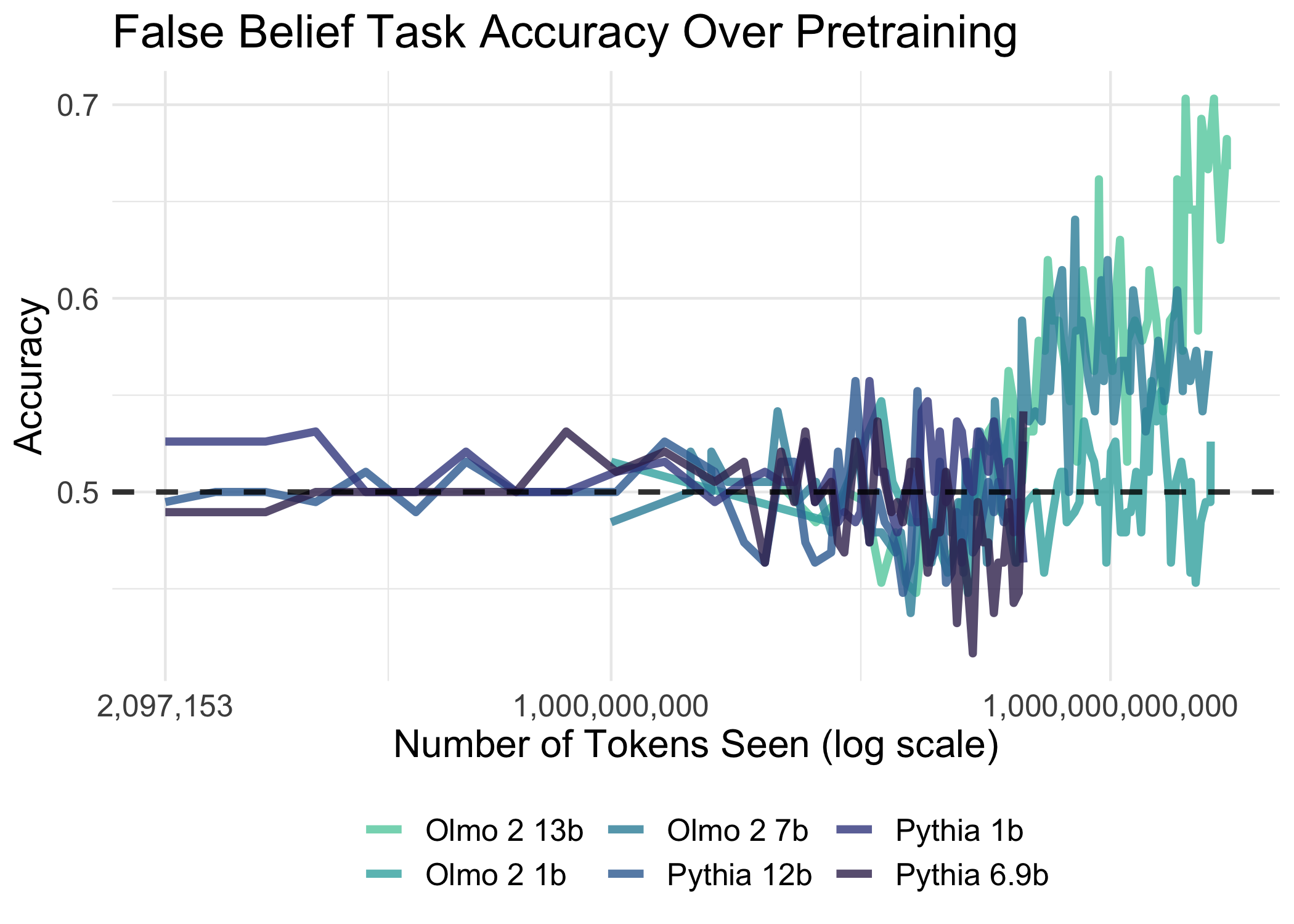}
  \caption{False belief task accuracy improvements over pretraining depend on model size and exposure to training tokens. The $0^{th}$ pretraining step has been removed and the x-axis begins with the next earliest selected step, to more clearly visualize performance at the more dynamic end of the range. Horizontal dashed line marks chance performance.}
  \label{fig:acc-by-tokens}
\end{figure}

\subsection{Language Model Selection}\label{sec:lm-selection}

We focused our study on the Olmo$2$ suite of openly-available LMs \citep{olmo20242}. In a replication of the False Belief (FB) task examining the final checkpoint of $41$ open-weight LMs, \citet{trott_language_2026}, Olmo$2$ $13b$ emerges as a top performer in this task. The Olmo$2$ suite is published along with multiple pretraining checkpoints, rendering Olmo$2$ LMs desirable substrates with which to examine learning dynamics.

We additionally explored the effect of LM size and volume of language exposure by charting the developmental trajectory of performance in approximately parameter-matched LMs from the Pythia suite \citep{biderman2023pythia}. Notably, these LMs were trained on an order of magnitude fewer tokens than their Olmo$2$ counterparts (\textbf{Table \ref{tab:lms}}).

All LMs were loaded and run using the HuggingFace \texttt{transformers} package \citep{wolf-etal-2020-transformers} on remote Lambda On-Demand instances \footnote{\url{https://docs.lambda.ai/public-cloud/on-demand/}}, selected according to availability at runtime (B200, H100 SXM, GH200, A100 SXM). We sampled log-spaced pretraining checkpoint indices within each pretraining stage (if there were multiple, as was the case for Olmo$2$), along with the stage's final checkpoint. For Olmo$2$, we also analyzed behavior at subsequent post-training stages. See Appendix \ref{sec:appendix} for a full list of LM checkpoints we examined.

\subsection{Procedure}\label{sec:lm-procedure}

We tokenized passage stimuli with the corresponding tokenizer for a given LM, then obtained the LM probability over Start and End tokens as likely passage completions. We used this to compute our critical measure, the log-odds of start versus end locations: $log_2(\frac{p(\textsc{start})}{p(\textsc{end})})$. We derived accuracy by considering a log-odds $>0$ for False Belief, and a log-odds $<=0$ in True Belief conditions as correct responses. Any other mapping was marked incorrect. Accuracy then consists of the proportion of log-odds marked correct.

For Situation Modeling tasks, we obtained the the probability over the correct and distractor token completions and computed the corresponding log-odds: $log_2(\frac{p(\textsc{correct})}{p(\textsc{distractor})})$. Consequently, a log-odds $>0$ was marked correct, and $<=0$ incorrect, from which we then computed accuracy. All scripts and data are available as GitHub\footnote{\url{https://github.com/seantrott/dev_tom/tree/main}} and Open Science Framework\footnote{\url{https://osf.io/ncypd/overview}} repositories. 

\begin{figure}[t]
  \includegraphics[width=\columnwidth]{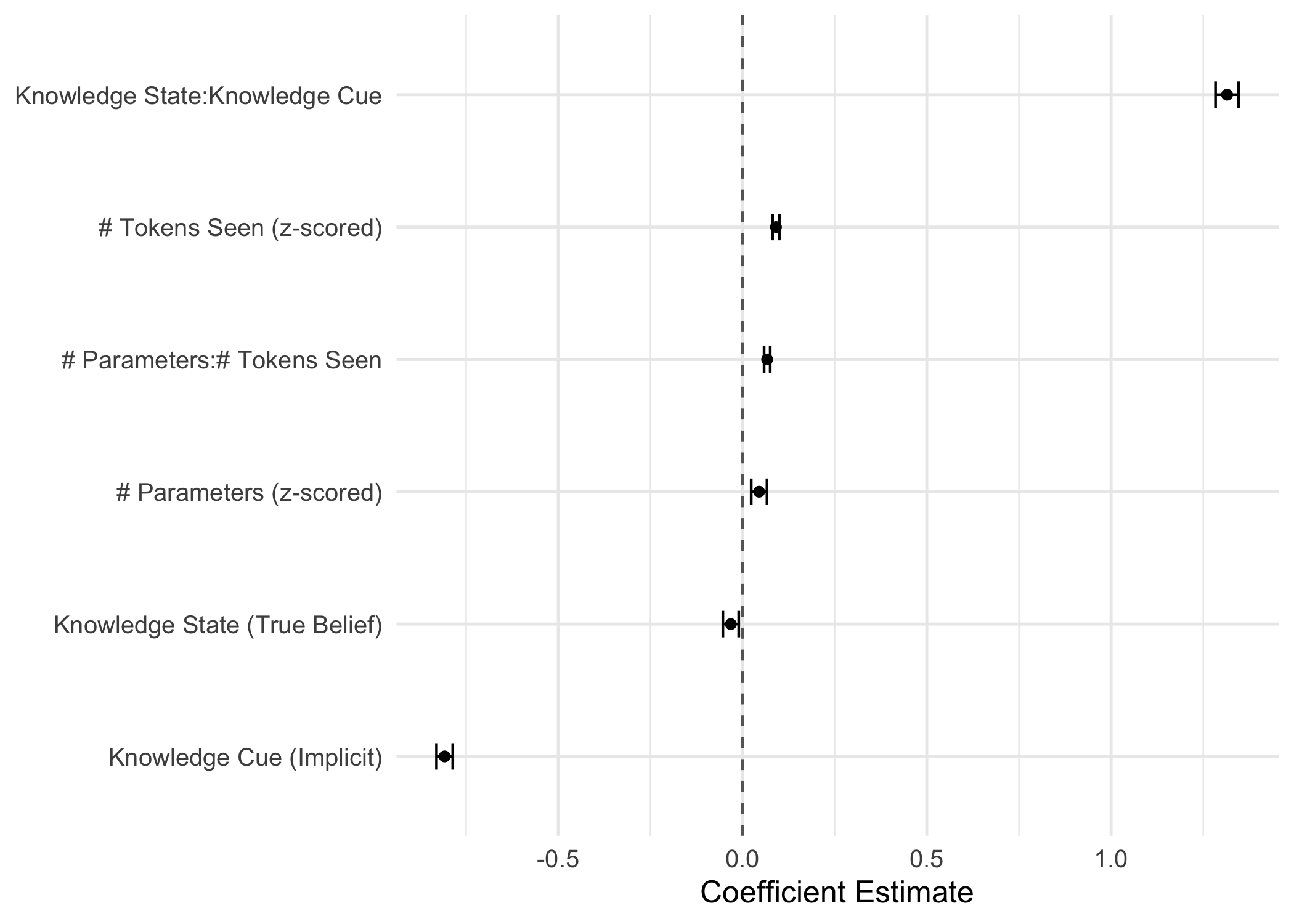}
  \caption{Coefficients of a linear mixed effects model predicting False Belief task accuracy from stimulus properties (Knowledge State, Knowledge Cue) and language model (LM) properties (z-scored number of tokens seen and z-scored number of LM parameters), as well as the corresponding interactions. LMs were more accurate overall for passages where the target agent should know where the item is to be found.}
  \label{fig:coefficients}
\end{figure}

\section{Results}

All analyses and visualizations were carried out in R, with mixed effects models  built using the \texttt{lme4} package \citep{bates2015fitting} and visualizations with \texttt{ggplot} \citep{wickham-ggplot}. 

\begin{figure}[t]
  \includegraphics[width=\columnwidth]{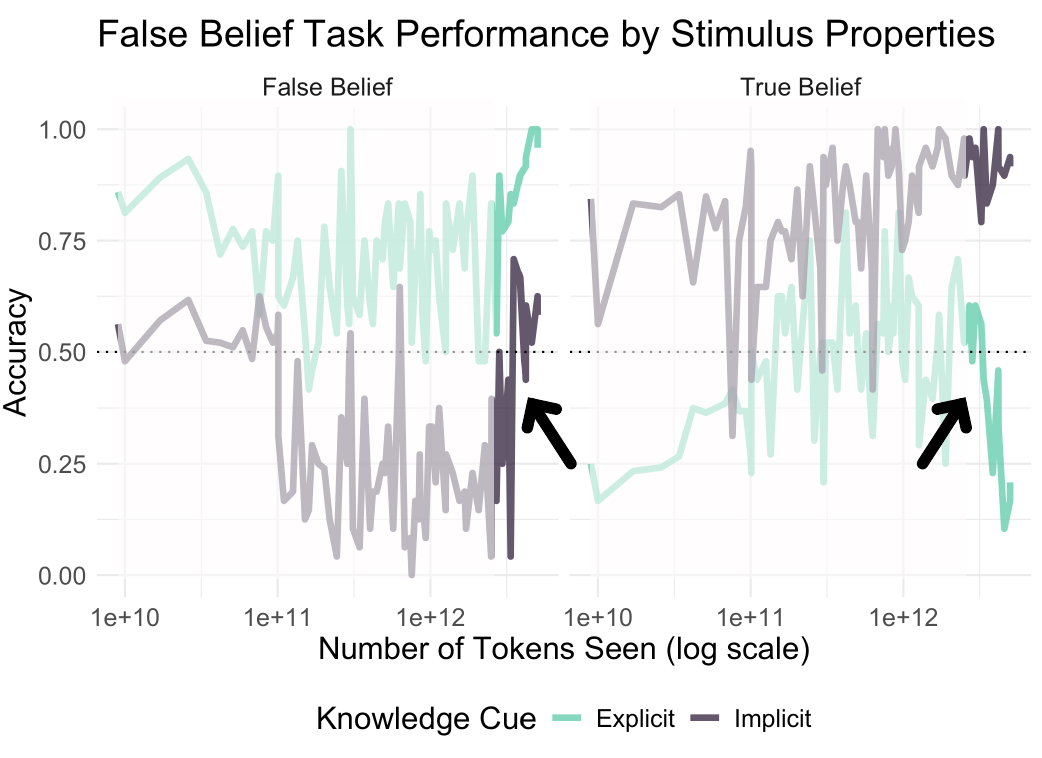}
  \caption{Olmo$2$ $13b$'s sensitivity for False Belief stimulus properties changes over the course of pretraining. Black arrows mark chance-level crossings in accuracy towards the end of pretraining, with sudden drops in accuracy for True Belief, Explicitly cued passages that coincide with sudden increases in accuracy for False Belief, Implicit passages. See also companion Appendix \textbf{Figure \ref{fig:olmo13b-logodds}}.}
  \label{fig:olmo13b-acc-by-stimproperties}
\end{figure}

\subsection{Pretraining Dynamics of Mentalizing}\label{sec:mentalizing-results}

We computed accuracy over token completions for False Belief (FB) Task passages \citep{trott_large_2023}, across select LMs from the Olmo$2$ \citep{olmo20242} and Pythia suites (\citet{biderman2023pythia}). We first evaluated whether LM FB task accuracy was systematically related to LM properties and stimulus features. We built a linear mixed effects model predicting binary correct/incorrect responses from  (z-scored) number of LM parameters as a proxy for model size, (z-scored) cumulative number of tokens seen, and the interaction of these two LM properties; as well as stimulus features such as (\texttt{Knowledge State}, the presence of \texttt{Knowledge Cue}), and their interaction. We additionally included random intercepts for \texttt{Language Model} and item \texttt{Start Location}. 

Unsurprisingly, LM model size predicted improvements in performance, but only in the presence of large volumes of training data: a significant positive interaction between number of parameters and number of tokens seen ($\beta = 0.07, SE = 0.008, p < 0.001$) indicated that model capacity alone, if under-trained, is insufficient to produce behavior consistent with mentalizing (\textbf{Figure \ref{fig:acc-by-model-size}}). This pattern is even clearer when visualizing pretraining dynamics: the under-trained Pythia LMs perform consistently at chance regardless of size; while the smallest member of the Olmo$2$ suite is unable to leverage the large amount of training tokens it has seen. Only Olmo$2$ $7b$ and $13b$, combining superior model size with greater language exposure, improve in accuracy toward the latter end of stage$1$ pretraining (\textbf{Figure \ref{fig:acc-by-tokens}}). 

\begin{figure}[t]
  \includegraphics[width=\columnwidth]{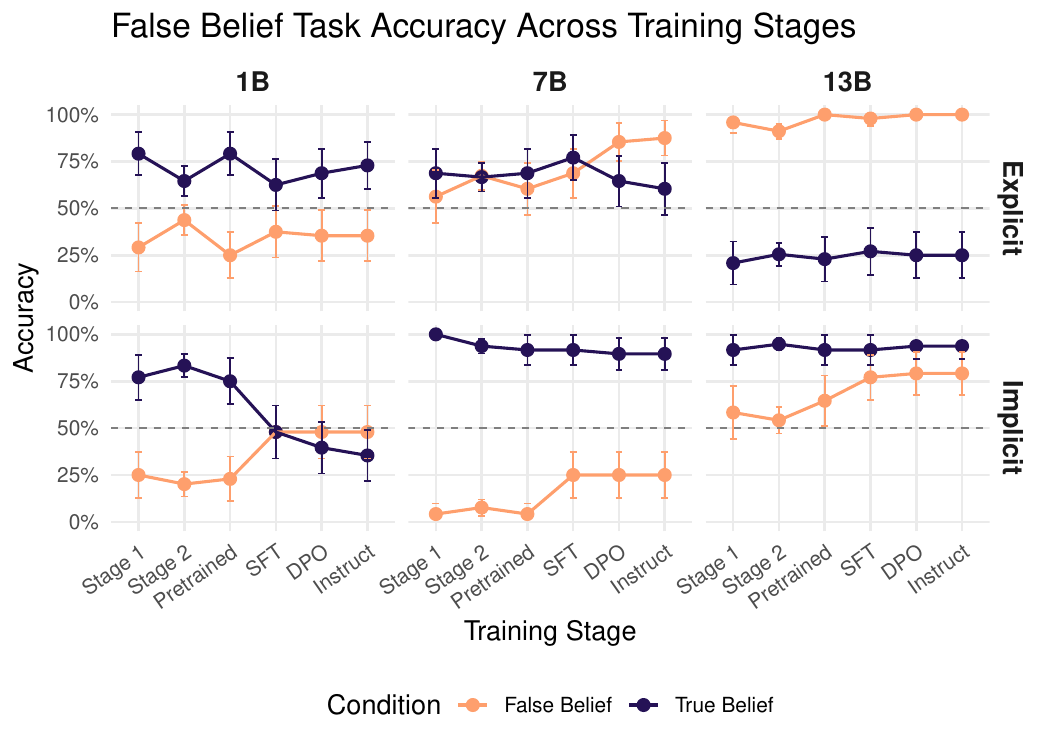}
  \caption{Olmo$2$ $13b$ post-training interventions improve performance under stimulus properties that are typically used to index mentalizing (False Belief, Implicit), but remains relatively uniform for all others, including low-accuracy ones that should only reflect the ability to report on factual properties of a scene (True Belief, Explicit). Subpanels depict Olmo$2$ $13b$ FB task accuracy across various training stages, colored according to \texttt{Knowledge State} condition, and rows indicating the presence or absence of \texttt{Knowledge Cue} within passages. See Appendix \textbf{Figure \ref{fig:all-lms-post-train-log-odds}} for the corresponding log-odds.}
  \label{fig:acc-post-train}
\end{figure}

Stimulus properties, such as \texttt{Knowledge State} and \texttt{Knowledge Cue} were also strong predictors of performance. We found a significant interaction between these two stimulus properties ($\beta = 1.31, SE = 0.0312, p < 0.001$; \textbf{Figure \ref{fig:coefficients}}; \textbf{Table \ref{tab:fb-tb-by-knowlcue}}). In general, LMs tested here performed best in the True Belief \texttt{Knowledge State} condition (\textbf{Table \ref{tab:fb-tb}}), and this was particularly the case in the passages with Implicit \texttt{Knowledge Cue} (\textbf{Table \ref{tab:fb-tb-by-knowlcue}}). Only Olmo$2$ $13b$ and $7b$ performed above chance for passages within the crucial False Belief condition, but they only did so when the \texttt{Knowledge Cue} was Explicit in the query and performed systematically below chance when the \texttt{Knowledge Cue} was absent (\textbf{Table \ref{tab:fb-tb-by-knowlcue}}). In all, it appeared that LMs were sensitive to the presence of verbs that explicitly referenced the target agent's beliefs as to the item's location. Notably, this sensitivity to \texttt{Knowledge Cue} manifests quite late in the first stage of Olmo$2$ $13b$ pretraining (\textbf{Figure \ref{fig:olmo13b-acc-by-stimproperties}}).

Finally, we compared performance in the Olmo$2$ LMs across each stage of training, including Stage 1, Stage 2, the fully Pretrained model, supervised fine-tuning (SFT), direct policy optimization (DPO), and instruction-tuning (Instruct). As \textbf{Figure \ref{fig:acc-post-train}} makes clear, accuracy was remarkably stable in 13$b$ across stages when the \texttt{Knowledge Cue} was Explicit; however, we did observe marked improvement (for 13$b$) in the False Belief condition when \texttt{Knowledge Cue} was Implicit, consistent with concurrent work \citep{kouwenhoven2026social}. See Appendix \textbf{Figure \ref{fig:all-lms-post-train-log-odds}} for a visualization of the changes in the corresponding log-odds across models and training stages. 
 
\begin{figure}[t]
  \includegraphics[width=\linewidth]{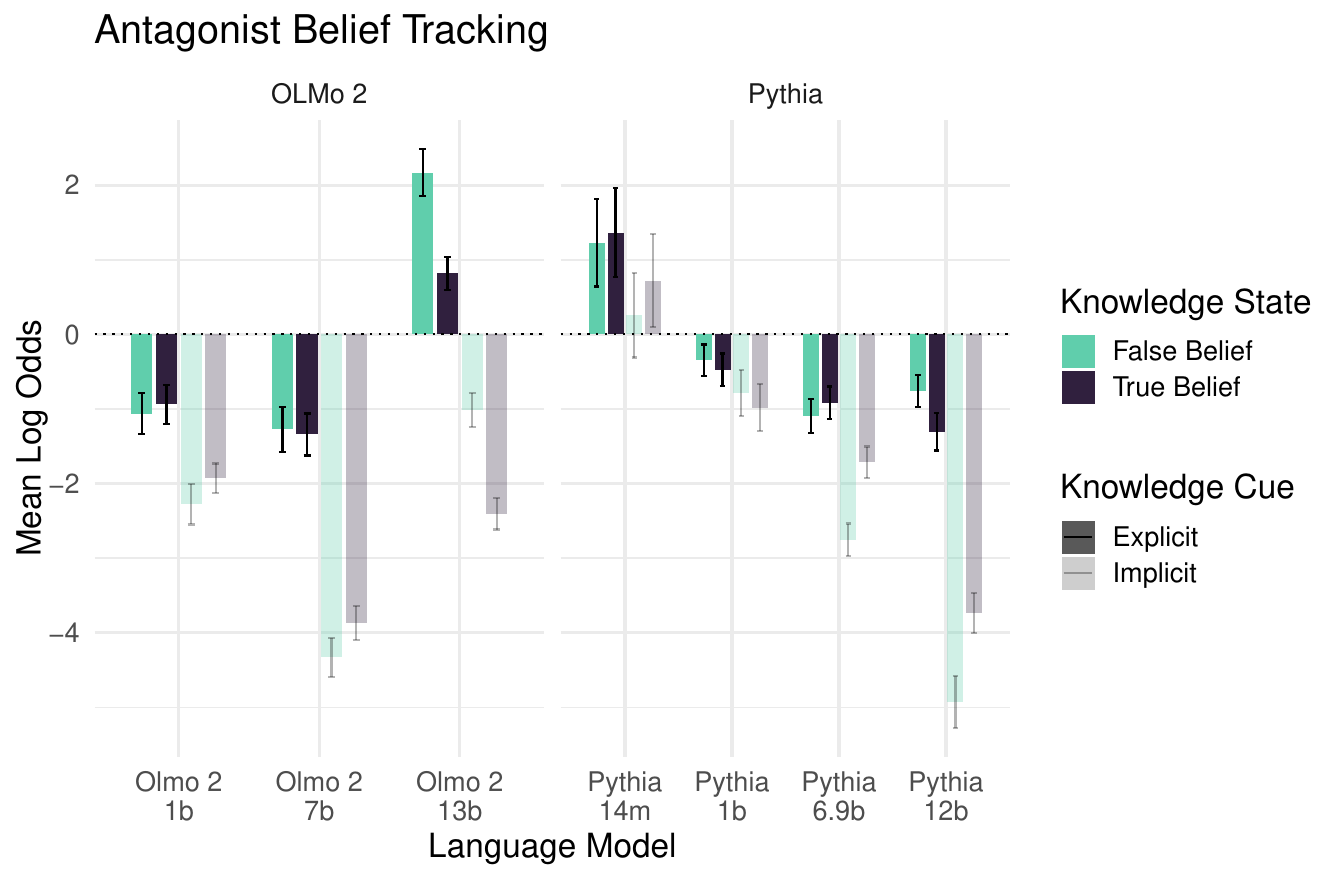} \hfill
  \caption {Antagonist Belief task mean log odds at the final checkpoint for all LMs tested (from \textbf{stage$1$} for the Olmo$2$ suite). Positive values indicate larger probabilities for Start locations; negative values indicate larger probabilities assigned to End locations. In this task, the end location is always correct; positive values index low-accuracy performance. Note: Antagonists cannot hold False Beliefs in this task---but Olmo$2$ $13b$ \textit{still} incorrectly assigns larger probabilities to Start locations according to the \textit{Target} agent's \texttt{Knowledge State}. See Appendix \textbf{Figure \ref{fig:all-lms-stage1-control-task-acc-and-lo}} for developmental trajectories.}
  \label{fig:all-lms-control}
\end{figure}

\subsection{Charting the Emergence of Situation Modeling}\label{subsec:sm_results}

In the work that follows, we focus our interpretation on Olmo$2$ $13b$, the LM that exhibited performance most consistent with mentalizing (\textbf{Figure \ref{fig:acc-by-model-size}}). Notably, the passages where Olmo$2$ $13b$ displayed late-stage, sudden drops in accuracy are those that (in principle) should require deploying a less complex cognitive ``toolkit''. In these True Belief, Explicit passages, the Target agent is privy to all the same information as the omniscient observer (the LM), and as a result, mentalizing itself is not strictly \textit{required} to emit correct responses. This points to an over-reliance on certain surface properties of the stimulus, such as the presence of non-factive verbs like ``thinks'' \citep{trott_large_2023, trott_language_2026, kouwenhoven2026social}). We conducted a series of ``stress-tests'' to further contextualize this finding, and probe the model's ability to report on factual properties of the scene by presenting it with modified queries (as described in Section \ref{sec:sm-stims}; \citet{shapira_clever_2024,riviere2025start}).

\subsubsection{Stress-Test: Tracking Antagonist Beliefs}\label{subsec:antagonist_beliefs}

The Antagonist agent in each passage always moved the item to its end location, in either the Target agent's presence (True Belief) or absence (False Belief). Consequently, LMs capable of constructing and maintaining a coherent Situation Model should assign larger probabilities to the item's End location relative to the Start when the query asks where the Antagonist is likely to search. At the final checkpoint, we found that Olmo$2$ $13b$ mistakenly assigned larger probabilities to Start locations in passage queries that explicitly referenced the Antagonist's beliefs (e.g. ``Marta \textbf{thinks}...''), an error pattern that only Pythia $14m$---orders of magnitude smaller---shared (\textbf{Figure \ref{fig:all-lms-control}}; Appendix \textbf{Figure \ref{fig:all-lms-stage1-control-task-acc-and-lo}}). In these cases, it appeared that the presence of the Explicit cue biased the LM to predict that the Antagonist would look for the item in the incorrect location. Moreover, even though the Antagonist could only have True Beliefs, Olmo$2$ $13b$ assigned larger probabilities to Start locations when passages contained Target agents with False Beliefs, even though their beliefs were not the ones the LM was expected to infer. 

We built a linear mixed effects model predicting Olmo$2$ $13b$ log-odds from \texttt{Knowledge State}, \texttt{Knowledge Cue}, their interaction, and the order in which locations were mentioned, which were counterbalanced across passages (Section \ref{sec:methods}), and a random intercept for start location. We found significant main effects of \texttt{Knowledge State} ($\beta = -1.35$, SE = $0.247$, $p < 0.001$, reference level: True Belief) , and \texttt{Knowledge Cue} ($\beta = -3.19$, SE = $0.247$, $p < 0.001$, reference level: Implicit), further reinforcing the notion that Olmo$2$ $13b$ suffers particular weaknesses with respect to conditions in which the agent in question should know exactly where to find the item.

\subsubsection{Stress-Test: Tracking Causal Structure}

Given Olmo$2$ $13b$'s failures to infer the Antagonist's most likely search location, we explored the LM's general capacity for modeling the events of the situation, such as reporting on where the item was located at the Start and End, and which agent moved the item to a given location (see Section \ref{sec:sm-stims} for a sample list of Situation Modeling queries). 

\begin{figure*}[t]
  \includegraphics[width=0.48\linewidth]{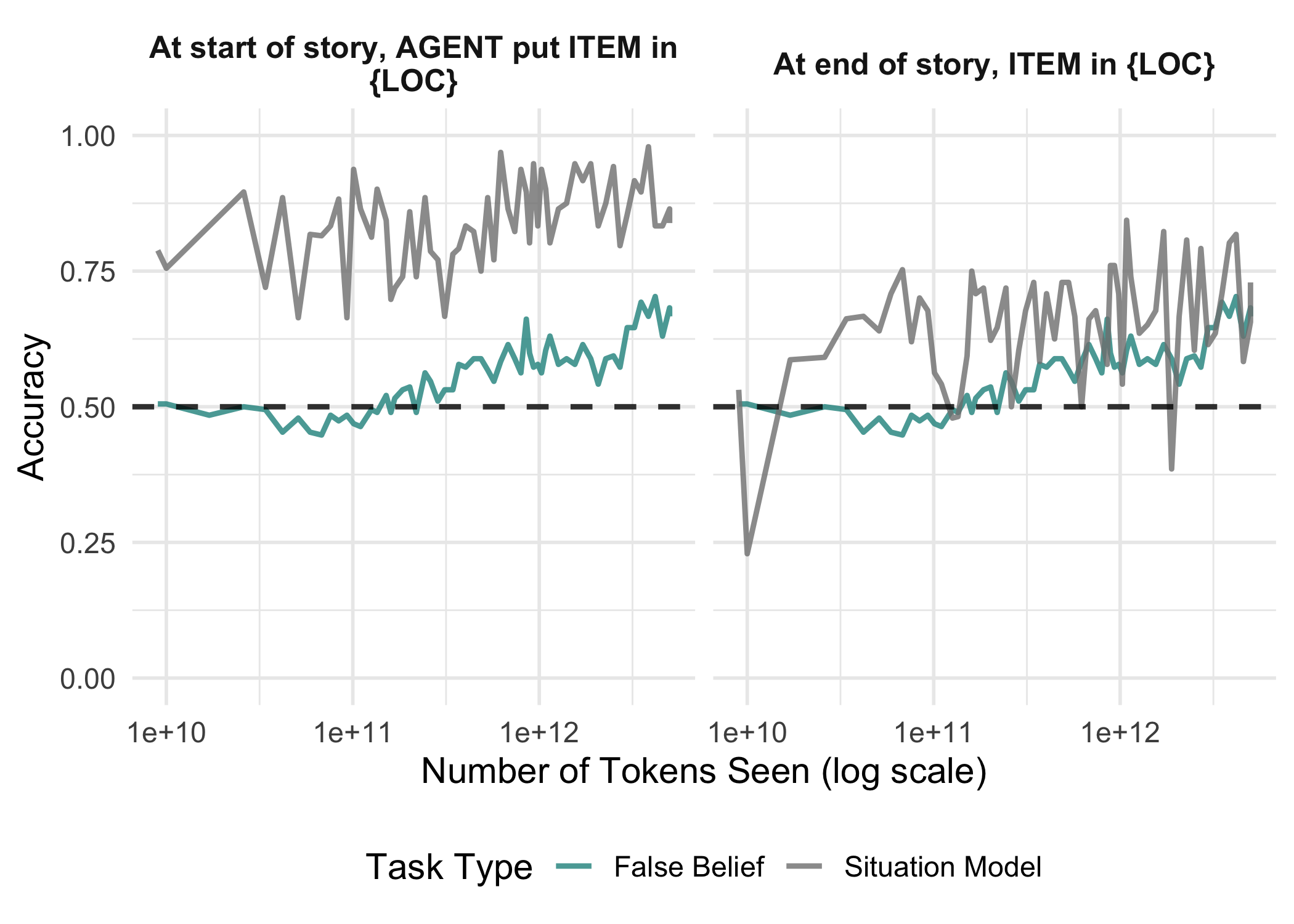} \hfill
  \includegraphics[width=0.48\linewidth]{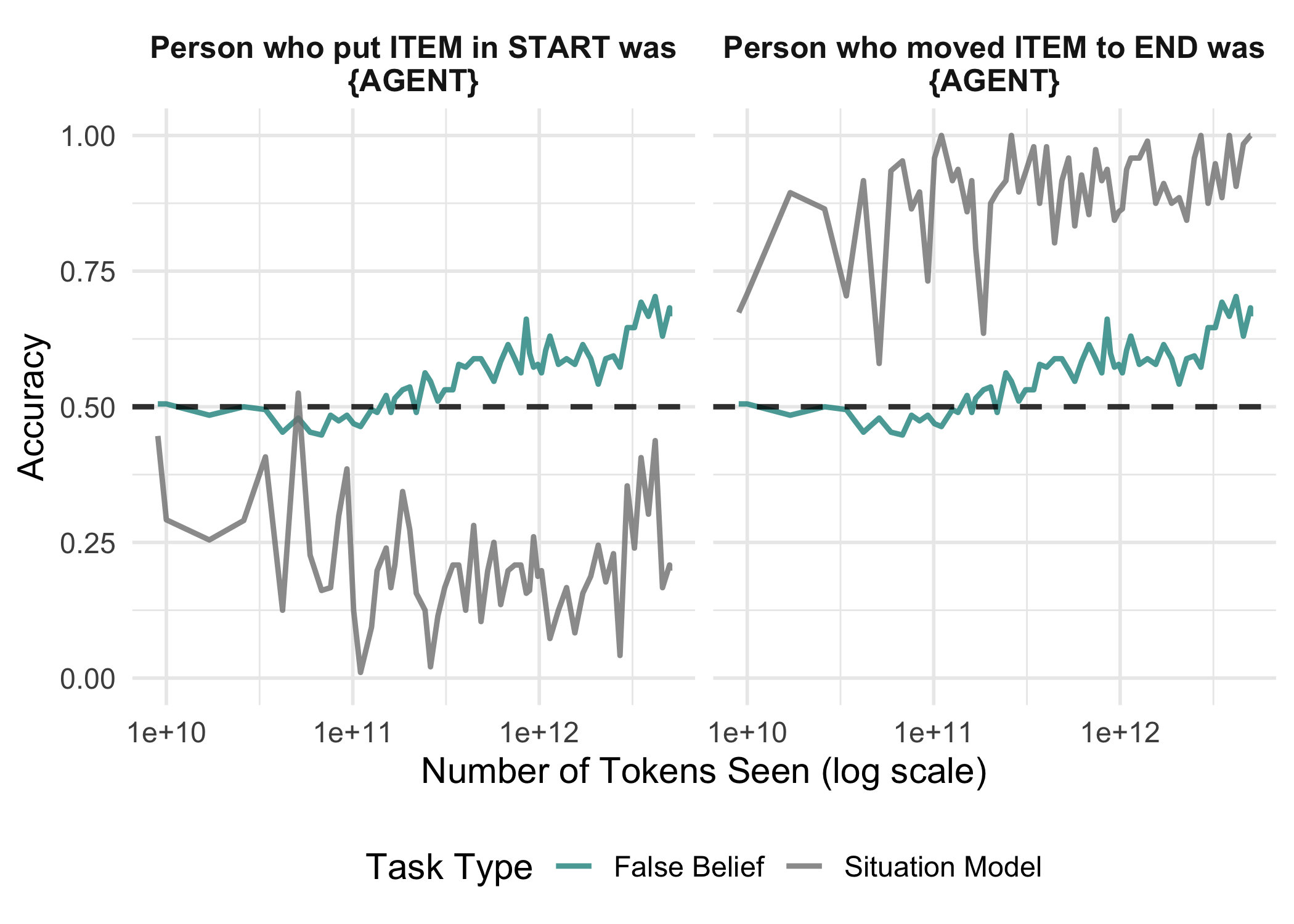}
  \caption {Olmo $2$ $13b$ situation model accuracy generally increases over the course of pretraining, its improvements \textit{precede} and generally \textit{exceed} False Belief task accuracies, but biases also emerge, particularly when the model is asked to track agents. Performance trajectories are colored according to Task Type, with False Belief accuracies in teal, and Situation Model accuracies in grey. Dashed horizontal line indicates chance performance.}
  \label{fig:sit-mod}
\end{figure*}

We found that improvements in Olmo$2$ $13b$'s Situation Model (SM) task performance generally \textit{preceded} and \textit{exceeded} its improvements in the False Belief (FB) task (\textbf{Figure \ref{fig:sit-mod}}). This was particularly true for questions about the location of an item, i.e., at the \textit{start} or \textit{end} of a passage; it was also true for questions about the identity of the agent who \textit{moved} the item between locations (i.e. the Antagonist), which is particularly striking given Olmo$2$ $13b$'s failures to accurately track the Antagonist's \textit{beliefs} when asked (\textbf{Figure \ref{fig:all-lms-control}}).

Conversely, Olmo$2$ $13b$ was unable to accurately report on which agent had placed the item in its start location (i.e. the Target agent), the agent whose \textit{beliefs} it was best able to track, most notably under False Belief conditions crucial to detecting the (in-principle) more complex mentalizing ability (\textbf{Figure \ref{fig:olmo13b-acc-by-stimproperties},\ref{fig:olmo13b-logodds}}). In fact, Olmo$2$ $13b$ displayed a remarkably consistent bias over the course of pretraining to emit larger probabilities for the ``Antagonist'' agent rather than the Target agent. 

Together, these results suggest that Olmo$2$ $13b$'s representation of the situation is somehow \textit{incoherent}: it can report the location of objects, and the Target agent's beliefs about those object locations (FBT), while struggling to report the identity of the character who originally moved the object, as well as the Antagonist's beliefs. In principle, performing the FBT correctly should depend on knowing which characters moved the item to each location, given that this is the crucial differentiating factor between their respective knowledge states. 

   
   

\section{Discussion}\label{sec:discussion}

This work adopted a developmental perspective to characterize the mentalizing and situation modeling capabilities of language models (LMs) across two broad ``families'': the Olmo$2$ $13b$ \citep{olmo20242} and Pythia suites \citep{biderman2023pythia}. We found that performance on the False Belief task (FBT) depended on both model size and training volume, and was also strongly influenced by stimulus properties---such as the use of a non-factive verb like ``thinks'' in the query to the LM (\textbf{Figures \ref{fig:acc-by-model-size}-\ref{fig:coefficients}; Tables \ref{tab:fb-tb}, \ref{tab:fb-tb-by-knowlcue})}. This latter finding was consistent with past work \citep{trott_language_2026, kouwenhoven2026social}, and notably, the pattern remained unchanged over the course of post-training, despite improvements in other conditions (\textbf{Figure \ref{fig:acc-post-train}}). 

In further stress-testing, we identified a similar effect of non-factive verbs in a task measuring a model's ability to accurately report the \textit{Antagonist's} belief states. This effect was particularly strong in cases where the \textit{Target} should hold a False Belief, suggesting the model experienced ``interference'' from the primary experimental manipulation (\textbf{Figure \ref{fig:all-lms-control}}). In a second block of stress-testing, we found that Olmo$2$'s ability to report an item's initial and final locations exceeded and preceded its ability to report second-order beliefs about those locations (\textbf{Figure \ref{fig:sit-mod}})---consistent with a relatively coherent situation model. Yet the model also displayed a systematic bias towards the Antagonist even in questions asking about the Target (i.e., the identity of the character who originally moved the item). Together with the first stress-test results, this paints a nuanced picture: situation model representations are surprisingly brittle along some dimensions, despite showing coherence in others.

These results also inform ongoing debates about the construct validity of mentalizing tests for LLMs \citep{hu_re-evaluating_2025}. In some respects, the developmental pattern is encouraging: situation modeling accuracy appears to precede improvements in mentalizing performance, consistent with the intuition that tracking the basic facts of a situation should precede---and perhaps support---reasoning about (false) beliefs about that situation. Yet this developmental coherence is complicated by incoherence along other dimensions, as described above. This could point to a collection of overlapping sensitivities rather than a unified mentalizing construct.

Finally, our results speak to longstanding debates about whether exposure to language is \textit{sufficient} for mentalizing-like behavior to emerge \citep{brown_why_1996, de_villiers_role_2014}. The results do support the conclusion that linguistic input can produce \textit{sensitivity} to belief states: FBT performance improved throughout training, and was strongest for large models with extensive linguistic exposure. At the same time, the first emergence of above-chance performance coincided with orders of magnitude more linguistic exposure than humans encounter in a lifetime---and this performance still displayed considerable fragility in the face of minor perturbations, consistent with other recent work \citep{shapira_clever_2024}. While we cannot generalize about LMs not observed in our current sample, these results do suggest that linguistic input is insufficient to produce robust, generalizable mentalizing behavior in \textit{these} LMs.

More broadly, the approach adopted here---tracing a behavior's developmental trajectory alongside likely preconditions and stress-tests---offers a general methodological template for evaluating LM capabilities. It joins a growing body of work assessing the ``ontogeny'' of LM behaviors and mechanisms \citep{chen_sudden_2023, tigges_llm_2024, riviere2025start}, as well as assessing the robustness of a particular behavior under various perturbations \citep{shapira_clever_2024}. This paradigm could complement existing benchmarking approaches and even serve as a method for construct validation.

\section*{Limitations}
The primary limitations of the present study stem from the stimulus set and the limited selection of LMs. We limited our stimuli to only those passages publicly accessible from \citet{trott_large_2023}, although many variants of the False Belief (FB) task exist to test mentalizing capabilities. It is possible that a distinct pattern of results would emerge for different variants of the FB task. Moreover, we constrained the analysis to LM suites that release pretraining checkpoints publicly. 


\section*{Acknowledgments}

We are grateful for insightful conversations with Tom Kouwenhoven, Michiel T. van der Meer, and  Max van Duijn, and to Stephen J. Hanson for generously providing access to an NVIDIA DGX-H200. 

\bibliography{custom, tom, dcv, generalizability}

\appendix

\section{Appendix}
\label{sec:appendix}


\subsection{False Belief Task: Sample Stimuli}\label{subsec:app-stims}

\begin{tcolorbox}

{\fontfamily{pcr}\selectfont
{\small \textbf{Knowledge State: False Belief}  \newline
\textbf{Knowledge Cue: Explicit} \newline
\textbf{Item Start Loc Appears First} \newline 
\textbf{Item End Loc Appears Last}\newline}

{\small David and Marta go out to get some wine for the party. When they get home, David stores the wine in the \textbf{garage} and grabs a drink from the fridge. Then, David goes out to get some snacks. \textbf{While David is gone}, Marta decides the wine would be best cooled, so she moves the wine out of the garage and into the \textbf{fridge}. David returns home and wants to put out the wine. David \textbf{thinks} the wine is in the }
}

\end{tcolorbox}
\begin{tcolorbox}

{\fontfamily{pcr}\selectfont
{\small \textbf{Knowledge State: False Belief}  \newline
\textbf{Knowledge Cue: Implicit} \newline
\textbf{Item Start Loc Appears First} \newline 
\textbf{Item End Loc Appears Last}\newline}

{\small David and Marta go out to get some wine for the party. When they get home, David stores the wine in the \textbf{garage} and grabs a drink from the fridge. Then, David goes out to get some snacks. \textbf{While David is gone}, Marta decides the wine would be best cooled, so she moves the wine out of the garage and into the \textbf{fridge}. David returns home and wants to put out the wine. David \textbf{goes} to get the wine from the }

}

\end{tcolorbox}
\begin{tcolorbox}

{\fontfamily{pcr}\selectfont
{\small \textbf{Knowledge State: True Belief}  \newline
\textbf{Knowledge Cue: Explicit} \newline
\textbf{Item Start Loc Appears First} \newline 
\textbf{Item End Loc Appears Last}\newline}

{\small David and Marta go out to get some wine for the party. When they get home, David stores the wine in the \textbf{garage} and grabs a drink from the fridge. However, Marta decides the wine would be best cooled. \textbf{David watches} Marta move the wine out of the garage and into the \textbf{fridge}. Then, David goes out to get some snacks. When he returns home, he wants to put out the wine. David \textbf{thinks} the wine is in the }

}
\end{tcolorbox}
\begin{tcolorbox}

{\fontfamily{pcr}\selectfont
{\small \textbf{Knowledge State: True Belief}  \newline
\textbf{Knowledge Cue: Implicit} \newline
\textbf{Item Start Loc Appears First} \newline 
\textbf{Item End Loc Appears Last}\newline}

{\small David and Marta go out to get some wine for the party. When they get home, David stores the wine in the \textbf{garage} and grabs a drink from the fridge. However, Marta decides the wine would be best cooled. \textbf{David watches} Marta move the wine out of the garage and into the \textbf{fridge}. Then, David goes out to get some snacks. When he returns home, he wants to put out the wine. David \textbf{goes} to get the wine from the }

}

\end{tcolorbox}

Variants of these four passages then cover the remaining permutations for the sequence of Start and End location mentions. We provide one example variant below, matched to the immediately preceding example box above in all but the sequence of start and end location mentions in the text: 
\begin{tcolorbox}

{\fontfamily{pcr}\selectfont
{\small \textbf{Knowledge State: True Belief}  \newline
\textbf{Knowledge Cue: Implicit} \newline
\textbf{Item End Loc Appears First} \newline 
\textbf{Item Start Loc Appears Last}\newline}

{\small David and Marta go out to get some wine for the party. When they get home, David grabs a drink from the \textbf{fridge} and stores the wine in the garage. However, Marta decides the wine would be best cooled. \textbf{David watches} Marta move the wine into the fridge from the \textbf{garage}. Then, David goes out to get some snacks. When he returns home, he wants to put out the wine. David \textbf{goes} to get the wine from the }
}
\end{tcolorbox}

\subsection{Situation Modeling: Minimal Pair Queries}
The original queries described in Section \ref{sec:sm-stims} were designed to match the way that information was presented in the original passage stimuli from \citet{trott_large_2023}. However, given Olmo$2$ $13b$'s sensitivity to surface properties of stimuli, it was possible that the results reported in \textbf{Figure \ref{fig:sit-mod}} may have been driven by the form of the queries, which varied in whether or not the agent was named, and in which verb was used (``put'', for Target; ``moved'' for Antagonist). To rule out this possibility, we presented modified Situation Modeling task queries that were minimally different from each other: 

\begin{itemize}
    \item At the start of the story, the book was in
    \item At the end of the story, the book was in
    \item The person who put the book in the box was
    \item The person who put the book in the basket was
\end{itemize}

We extracted probabilities over correct and distractor locations and agents from Olmo$2$ $13b$'s final checkpoint, and we found the results consistent with those obtained for the original Situation Modeling query set. Namely, the LM displayed a bias in favor of the Antagonist agent, even for queries for which the Target agent was the correct answer. 

\subsection{Additional Performance Visualizations}

We additionally include here LM performance visualized as the log of the ratios of probabilities assigned to the Start versus End locations, which offers a transparent view of the LM's ``responses''. All of the figures and data tables below are referenced in the Main Text, along with the figures and results they supplement.  

\begin{figure}[t]
  \includegraphics[width=\linewidth]{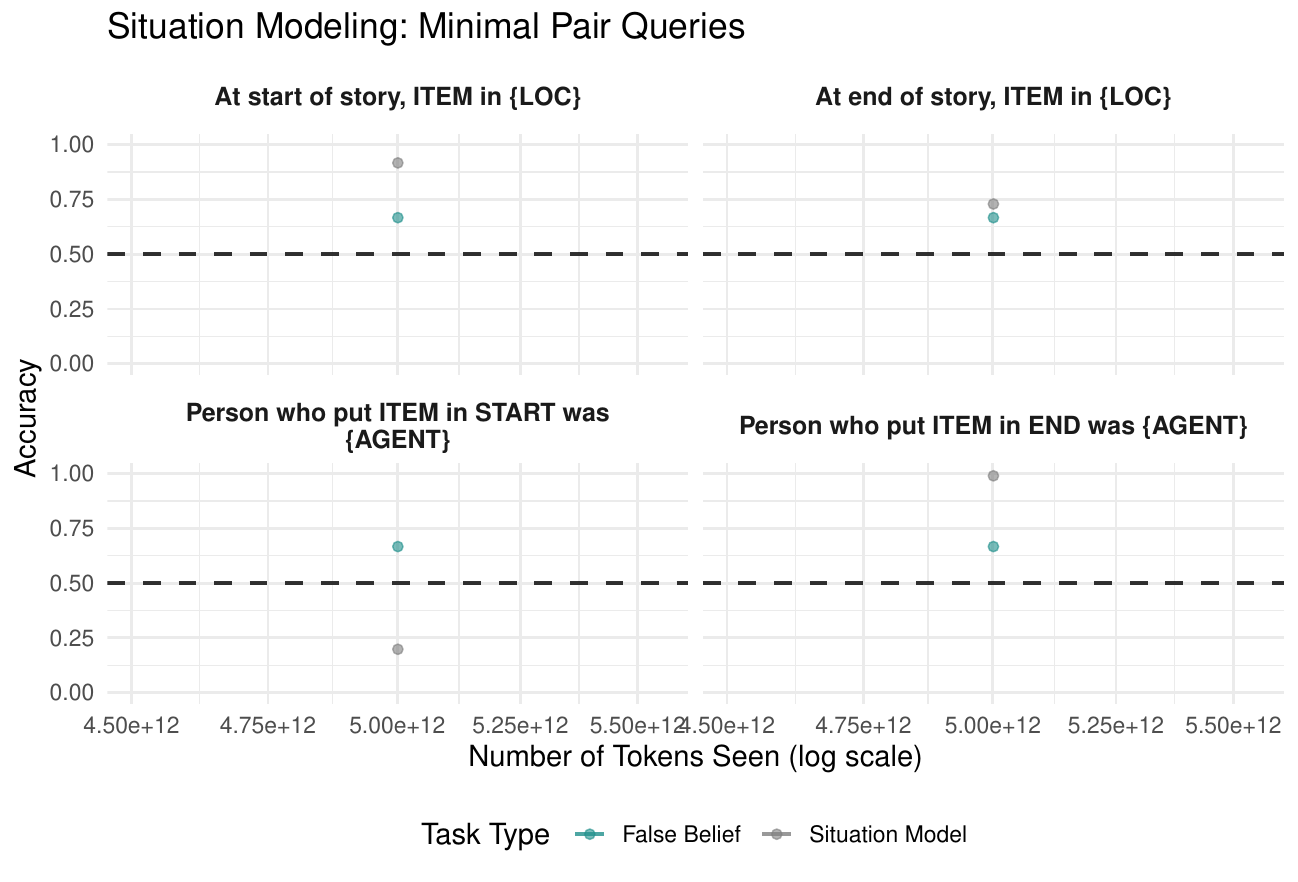} \hfill
  \caption {Companion to Main Text \textbf{Figure \ref{fig:sit-mod}}. Depicts Olmo$2$ $13b$'s final checkpoint accuracies for minimal pair versions of the original Situation Modeling set (subpanel titles illustrate the form of the new queries). Situation Modeling results are in grey; False Belief task performance for reference, in teal. These results match the overall trends observed in \textbf{Figure \ref{fig:sit-mod}}.}
  \label{fig:sm-min-pairs}
\end{figure}

\begin{table}
  \centering
  \begin{tabular}{lc}
    \hline
    \textbf{LM} & \textbf{MaxTokensSeen} \\
    \hline
    \verb|Pythia 14m|    &            \\
    \verb|Pythia 1b|     &    {~299B}       \\
    \verb|Pythia 6.9b|     &          \\
    \verb|Pythia 12b|     &            \\\hline
    \verb|Olmo 2 1b|      &  {4T}          \\
    \verb|Olmo 2 7b|     &  {~3.9T}          \\
    \verb|Olmo 2 13b|     & {5T}           \\\hline
  \end{tabular}
  
  \caption{Language models (LMs) examined in the present study, and the cumulative maximum number of tokens they observed by the end of pretraining. Olmo$2$ tokens are reported for ``stage 1'' of pretraining.}
  \label{tab:lms}
\end{table}

\begin{table}[h]
    \centering
    \begin{tabular}{lcc}
        \hline
        \textbf{Model} & \textbf{False Belief} & \textbf{True Belief} \\
        \hline
        \hline
        \texttt{Olmo 2 13b}  & 0.485 & \textbf{0.642} \\
        \texttt{Olmo 2 7b}   & 0.451 & \textbf{0.612} \\
        \texttt{Olmo 2 1b}  & 0.468 & \textbf{0.531} \\
        \texttt{Pythia 12b}  & 0.343 & \textbf{0.641} \\
        \texttt{Pythia 6.9b} & 0.395 & \textbf{0.585} \\
        \texttt{Pythia 1b}   & 0.491 & \textbf{0.527} \\
        \hline
    \end{tabular}
    \caption{Mean FB task accuracy by Knowledge State. Bolded entries correspond to the larger accuracy in each row. All LMs tested, regardless of family or size, performed best in the False Belief Task when the target agent had the same information as the observer (e.g. LMs perform egocentrically, on average). Note that Pythia $12b$ approximates Olmo$2$ $13b$'s accuracy on the True Belief, but not False Belief, condition.}
    \label{tab:fb-tb}
\end{table}

\begin{table*}[h]
    \centering
    \begin{tabular}{lcccc}
        \hline
        & \multicolumn{2}{c}{\textbf{False Belief}} & \multicolumn{2}{c}{\textbf{True Belief}} \\
        \textbf{Model} & \textbf{Explicit} & \textbf{Implicit} & \textbf{Explicit} & \textbf{Implicit} \\
        \hline
        \hline
        \texttt{Olmo 2 13b}  & 0.675 & 0.295 & 0.503 & \textbf{0.781} \\
        \texttt{Olmo 2 7b}   & 0.618 & 0.284 & 0.538     & \textbf{0.686}     \\
         \texttt{Olmo 2 1b}   & 0.519 & 0.416 & 0.516 & \textbf{0.546} \\
        \texttt{Pythia 12b}  & 0.399     & 0.288     & 0.602     & \textbf{0.681}     \\
        \texttt{Pythia 6.9b} & 0.425     & 0.365     & 0.573     & \textbf{0.596}     \\
        \texttt{Pythia 1b}   & 0.520     & 0.462     & 0.478     & \textbf{0.577}     \\
        \hline
    \end{tabular}
    \caption{Mean FB Task Accuracy Over Pretraining, by Knowledge State and Knowledge Cue. The crossover interaction (significant, according to the mixed effects model described in Results Section \ref{sec:mentalizing-results} between \texttt{Knowledge State} and \texttt{Knowledge Cue} is appreciable here. Note that these are mean accuracies over pretraining, not the final checkpoint. For Olmo$2$ $13b$, False Belief, Implicit mean accuracy is lower than the final checkpoint's accuracy, which achieves slightly greater-than-chance performance.}
    \label{tab:fb-tb-by-knowlcue}
\end{table*}

\begin{figure*}[t]
  \includegraphics[width=0.99\linewidth]{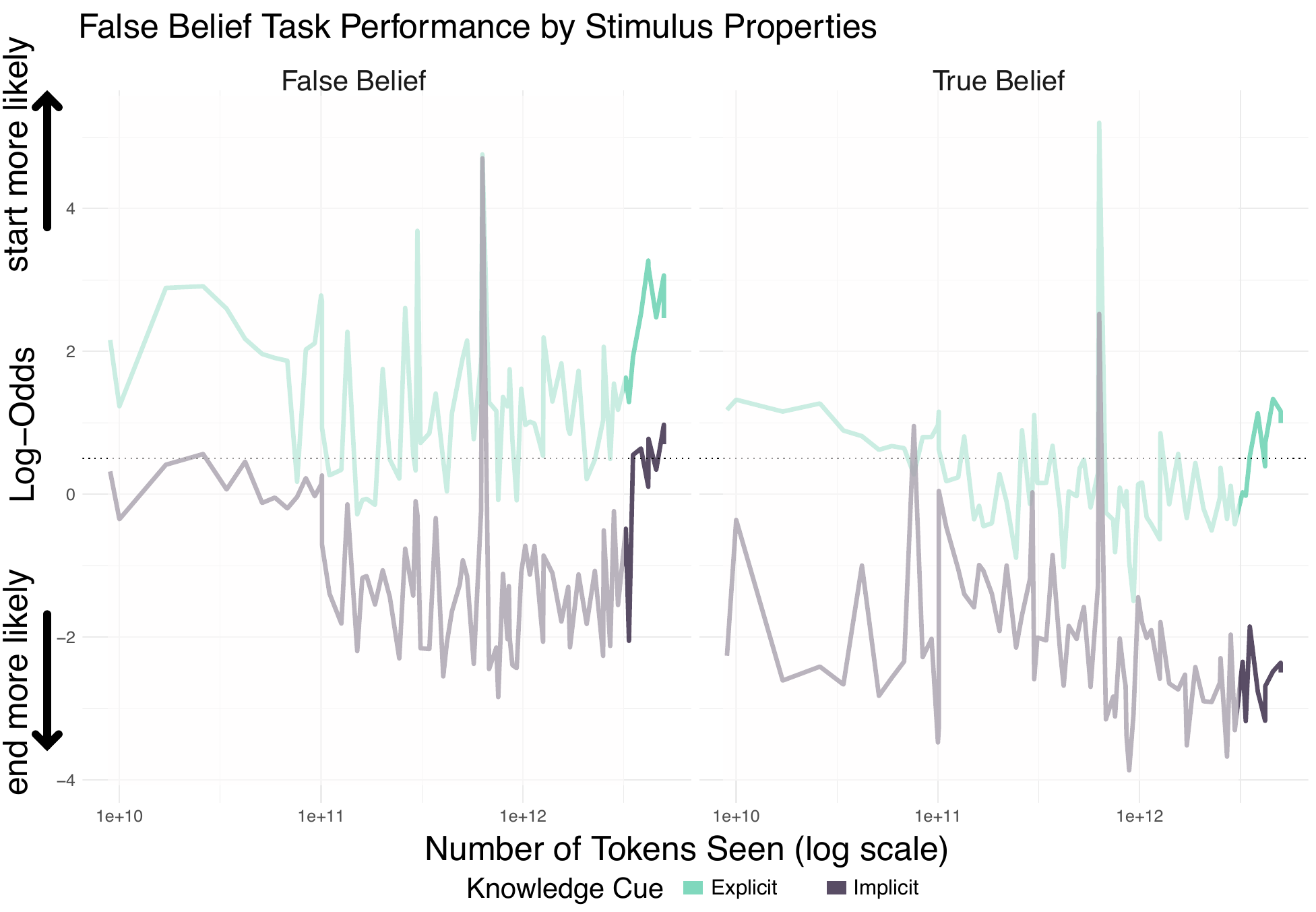} \hfill
  \caption {Companion to \textbf{Figure \ref{fig:olmo13b-acc-by-stimproperties}}. Log-odds of the ratio of probabilities assigned to Start versus End locations by the number of tokens observed during pretraining, and split by \texttt{Knowledge State} and color-coded by \texttt{Knowledge Cue}. We added an opacity layer over each subpanel to better highlight the differences in log-odds between the majority of stage$1$ pretraining and the sudden shifts that emerge towards the end.}
  \label{fig:olmo13b-logodds}
\end{figure*}

\begin{figure}[t]
  \includegraphics[width=\linewidth]{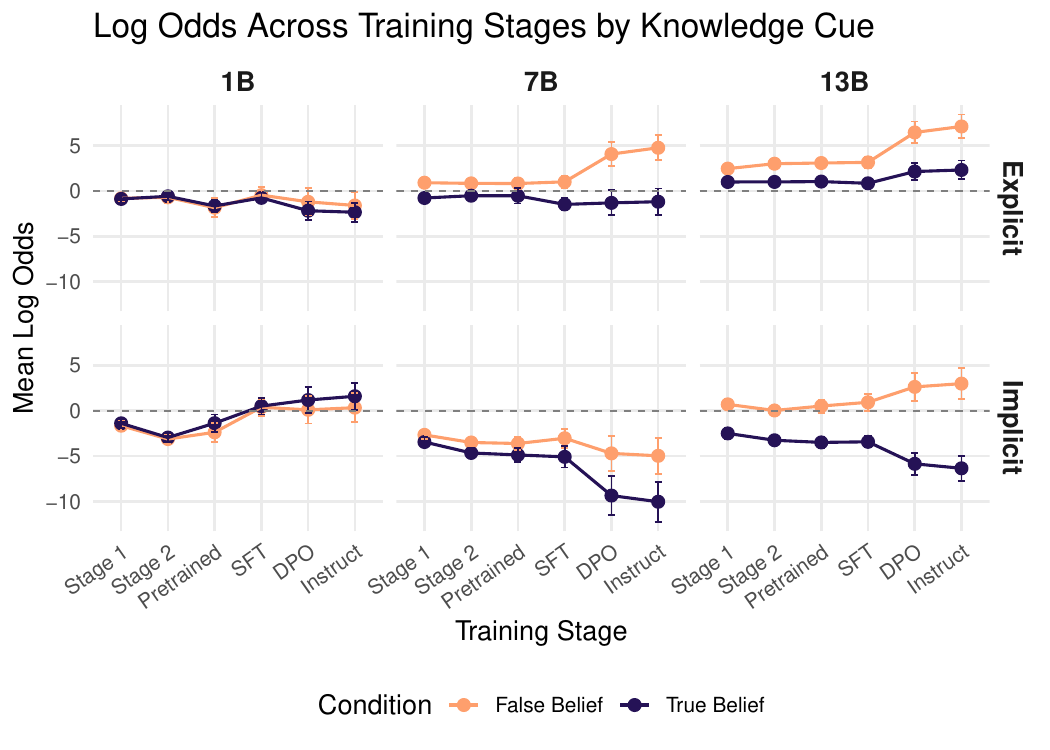} \hfill
  \caption {Companion to Main Text \textbf{Figure \ref{fig:acc-post-train}}. Subpanels depict mean log-odds across various training stages, colored according to \texttt{Knowledge State}, and with rows indicating whether the \texttt{Knowledge Cue} was Explicit or Implicit.}
  \label{fig:all-lms-post-train-log-odds}
\end{figure}

\begin{figure*}[t]
  \includegraphics[width=0.48\linewidth]{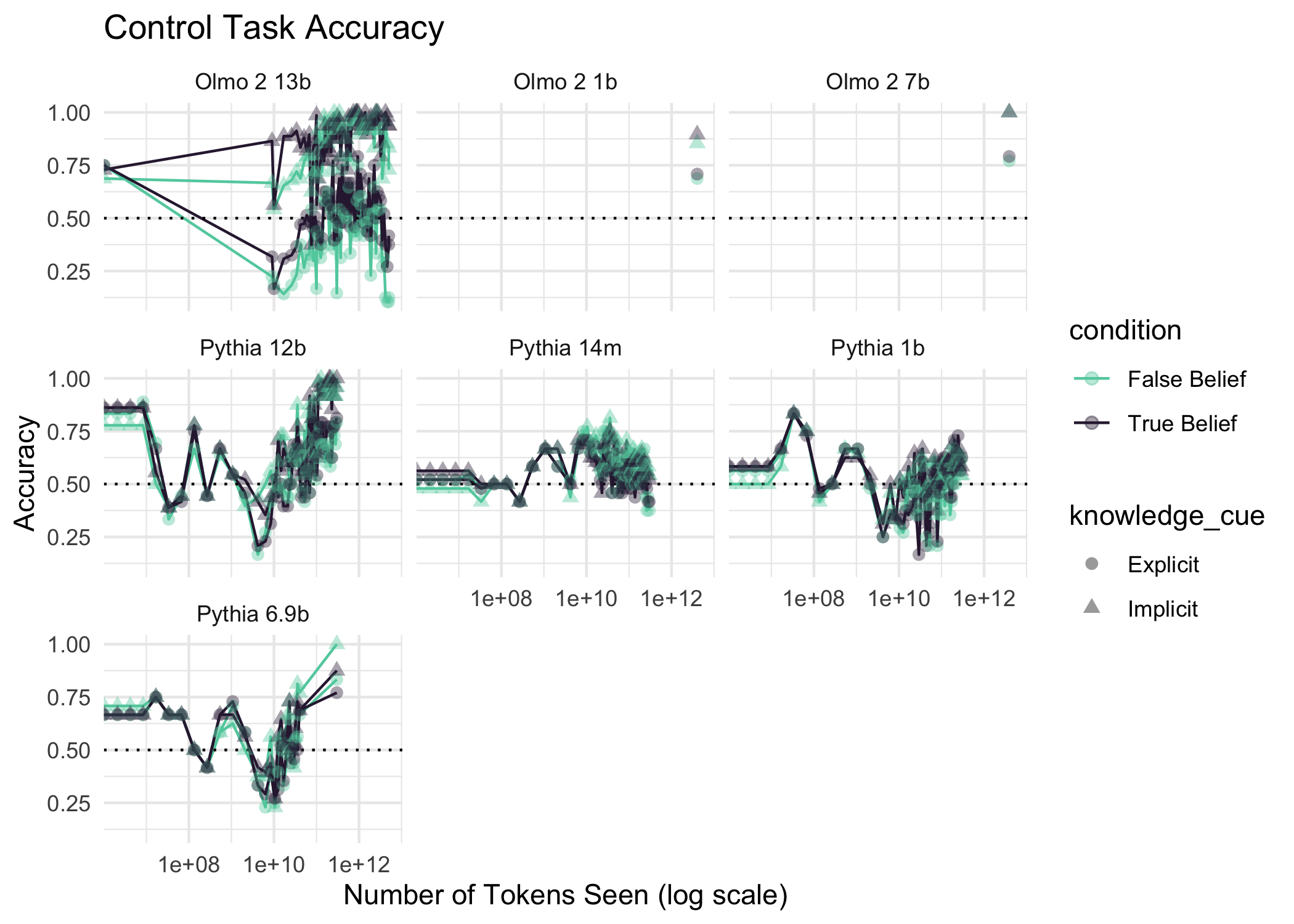} \hfill
  \includegraphics[width=0.48\linewidth]{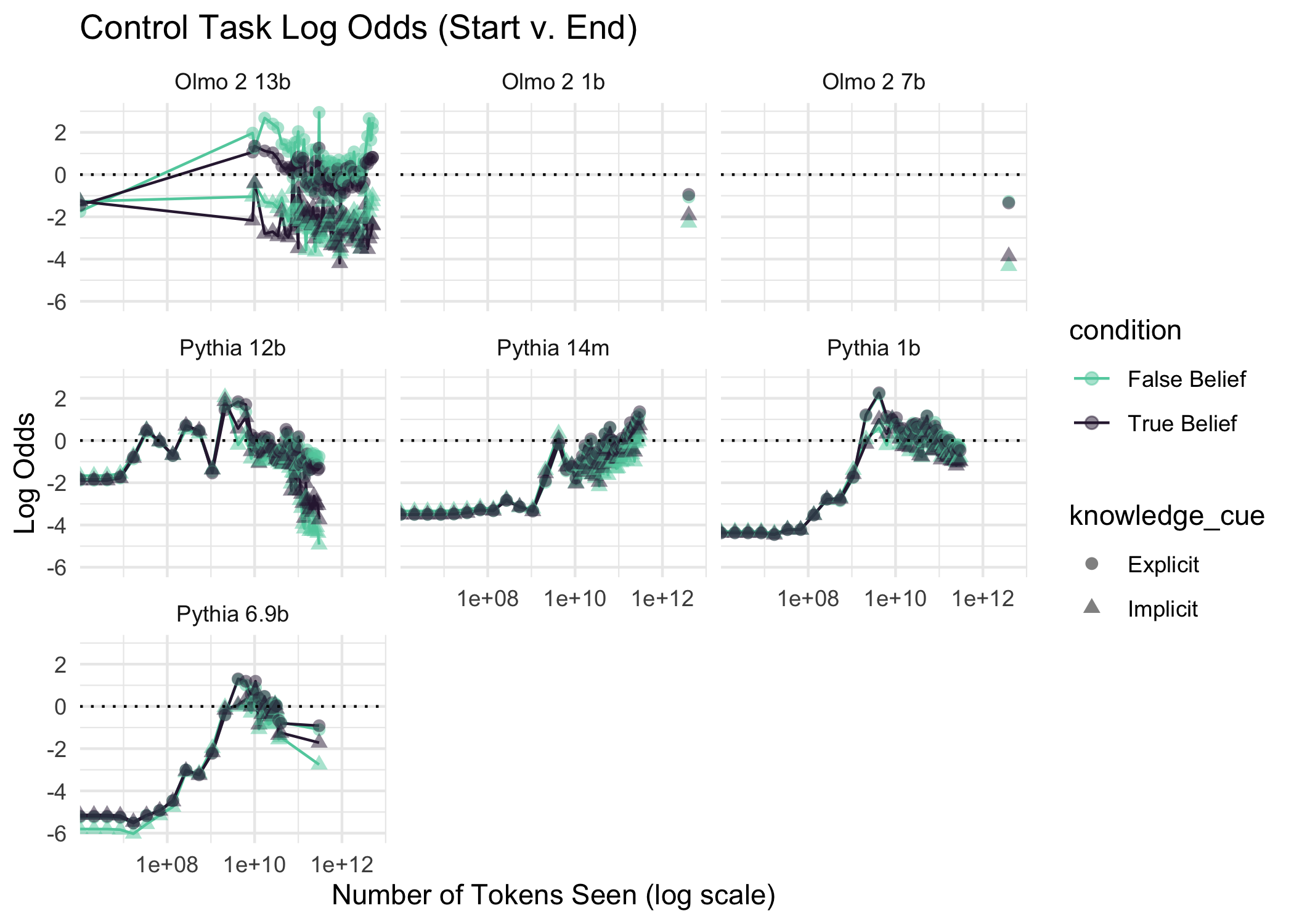} \hfill
  \caption {Control task (Antagonist Belief Tracking) accuracies for all LMs tested, for the Olmo$2$ suite LMs' final checkpoint of stage$1$; and over the course of pretraining for Pythia suite LMs and Olmo$2$ $13b$. Observations are color-coded by \texttt{Knowledge State} and the marker shape corresponds to the presence or absence of \texttt{Knowledge Cue} in the passages. Companion to Main Text \textbf{Figure \ref{fig:all-lms-control}}.}
  \label{fig:all-lms-stage1-control-task-acc-and-lo}
\end{figure*}

\end{document}